\pdfoutput=1

\documentclass[11pt]{article}

\usepackage{acl}

\usepackage{times}
\usepackage{latexsym}
\usepackage{graphicx}
\usepackage{float}

\usepackage[T1]{fontenc}

\usepackage[utf8]{inputenc}

\usepackage{microtype}

\usepackage{makecell}
\usepackage{booktabs}

\usepackage{subfig}
\usepackage{amsmath}
\usepackage{amsfonts}
\usepackage{multirow}
\usepackage{dblfloatfix}
\usepackage{scrextend}
\usepackage{url}

\title{Life after BERT: What do Other Muppets Understand about Language?}

\author{
    Vladislav Lialin\thanks{
        \hspace{0.5em}The first two authors made equal contribution to this work. Please direct correspondence to \texttt{vlialin@cs.uml.edu}, and \texttt{kevin\_zhao@uml.edu}.
    }
    \quad {\bf Kevin Zhao\footnotemark[1]}
    \quad {\bf Namrata Shivagunde}
    \quad {\bf Anna Rumshisky} \\
    Department of Computer Science \\ 
    University of Massachusetts Lowell \\
    \texttt{
    vlialin@cs.uml.edu,
    kevin\_zhao@uml.edu,}\\
    \texttt{nshivagu@cs.uml.edu,
    arum@cs.uml.edu} \\
}

\begin{document}
\maketitle

\begin{abstract}
Existing pre-trained transformer analysis works usually focus only on one or two model families at a time, overlooking the variability of the architecture and pre-training objectives.
In our work, we utilize the oLMpics benchmark and psycholinguistic probing datasets for a diverse set of 29 models including T5, BART, and ALBERT.
Additionally, we adapt the oLMpics zero-shot setup for autoregressive models and evaluate GPT networks of different sizes.
Our findings show that none of these models can resolve compositional questions in a zero-shot fashion, suggesting that this skill is not learnable using existing pre-training objectives.
Furthermore, we find that global model decisions such as architecture, directionality, size of the dataset, and pre-training objective are not predictive of a model's linguistic capabilities.
The code is available on GitHub \footnote{\href{https://github.com/kev-zhao/life-after-bert}{\texttt{github.com/kev-zhao/life-after-bert}}}.

\end{abstract}

\section{Introduction}


After the initial success of transfer learning in natural language processing \cite{Howard2018UniversalLM,Peters2018DeepCW}, the number of pre-trained models in NLP has increased dramatically \cite{Radford2018ImprovingLU,devlin2018bert,lewis2019bart,liu2019roberta,raffel2019exploring,lan2019albert,dong2019unified}. 
However, there is a limited understanding of why certain models perform better than others and what linguistic capabilities they acquire through pre-training.

While a lot of work has been done to evaluate these models on general natural language understanding datasets \cite{wang2018glue,wang2019superglue,lai2017race}, such datasets 
%
%
do not allow researchers to identify the specific linguistic capabilities of a model.
Furthermore
the performance on these datasets results from a combination of pre-trained knowledge and task-specific information learned through fine-tuning.

Probing tasks \cite{talmor2019olmpics,Zagoury2021WhatsTB,mccoy-etal-2019-right,Goldberg2019AssessingBS} give a promising solution to this problem, as they evaluate specific capabilities of pre-trained models, and in many cases, these tasks are designed for zero-shot evaluation, which reveals the knowledge that models have actually learned purely through the upstream task.
Currently, most in-depth analysis studies focus on one or two model families. Many analysis papers only probe $\mathrm{BERT}$ and similar models \cite{ettinger2020bert,kobayashi2020attention,Soler2020BertKnows,ravichander2020systematicity,Zagoury2021WhatsTB,Kassner2020ArePL,Mohebbi2021ExploringTR,Clark2020TransformersAS,Liu2021ProbingAT}. Fortunately, this trend is changing and now we see more papers that probe models such as $\mathrm{ALBERT}$, $\mathrm{T5}$ or $\mathrm{BART}$ \cite{Mosbach2020OnTI,Phang2021FineTunedTS,Jiang2021HowCW}. However, only a small number of analysis papers have probed multiple (three or more) model families \cite{Zhou2021RICAER,Ilharco2021ProbingCL}.

In our work, we test 8 families of models on oLMpics tasks \cite{talmor2019olmpics} and 6 families on psycholinguistic tasks from \citet{ettinger2020bert}. These models differ in size, architecture, pre-training objective, dataset size, and have other small yet important differences. Such a diverse set of models provides a broader view of what linguistic capabilities are affected by the change of any of these properties.
We also include several distilled models in our analysis.
We find that different models excel in different symbolic reasoning tasks, suggesting that \textit{slight differences related to optimization or masking strategy might be more important than the pre-training approach, dataset size, or architecture}.
Furthermore, in contrast to \citet{radford2019language}, we find that for oLMpics tasks, model size rarely correlates with the model performance.
In addition, we observe that all models fail on composition tasks when evaluated in a zero-shot fashion.

\section{Related Work}


Pre-trained model analysis is a rapidly growing area in NLP today. There exists a number of methods for analyzing internal representations of a model, 
including structured head and FCN pruning \cite{michel2019sixteen,voita2019analyzing,prasanna2020bert}, residual connection and LayerNormalization analysis \cite{kovaleva2021busters,kobayashi2021IncorporatingRA}, and analyzing attention patterns \cite{clark2019does,kovaleva2019revealing}.

Compared to these methods, probing tasks \cite{Conneau2018WhatYC,tenney2019bert} provide a more direct way to evaluate what a model can and cannot accomplish.
While it is possible to probe embeddings or hidden representations directly \cite{tenney2019bert,liu2019linguistic}, the adoption of pre-trained language models has made it possible to evaluate such models by framing probing tasks close to the original model objective \cite{radford2019language,talmor2019olmpics,ettinger2020bert,Goldberg2019AssessingBS}.

However, when a research area moves this quickly, it can be hard to keep up with many new models. Most of the existing research \cite{Soler2020BertKnows,Zagoury2021WhatsTB,Kassner2020ArePL} papers compare only one or two model families. Even some of the most recent works only probe $\mathrm{BERT}$ or very similar models \cite{Zagoury2021WhatsTB,Liu2021ProbingAT}.
Only a small number of analysis papers have probed multiple (three or more) model families \cite{Zhou2021RICAER,Ilharco2021ProbingCL}.

In contrast to existing work, we perform a large-scale probing of 29 models across 8 different model families.
We apply the existing probing benchmarks, namely, oLMpics \cite{talmor2019olmpics} and psycholinguistic datasets \cite{ettinger2020bert}, to models that differ in the pre-training objective, datasets, size, architecture, and directionality.

\begin{table*}
\centering
\small
\begin{tabular}{ l|c c c c c c} 
\toprule
Model & Parameters & Pre-training Data Size & Enc-Dec & Autoregressive & Tokenization & Vocab. Size \\
\midrule
    $\mathrm{BERT}$ & 110M - 340M & 16 GB & No & No & WordPiece & 30,522 \\
    $\mathrm{RoBERTa}$ & 355M & 160 GB & No & No & BPE & 50,265\\
    $\mathrm{DistilBERT}$ & 66M & 16 GB & No & No & WordPiece & 30,522 \\
    $\mathrm{AlBERT}$ & 12M - 235M & 16 GB & No & No & SentencePiece & 30,000 \\
    $\mathrm{GPT2}$ & 124M - 1.5B & 40GB\footnotemark & No & Yes & BPE & 50,257 \\
    $\mathrm{UniLM}$ & 340M & 16 GB & No & N/A & WordPiece & 28,996 \\
    $\mathrm{BART}$ & 406M & 160 GB & Yes & Yes & BPE & 50,265 \\
    $\mathrm{T5}$ & 223M-2.8B & 750 GB & Yes & Yes &  SentencePiece & 32,128 \\
\bottomrule
\end{tabular}
\caption{Model families used in this study. Enc-Dec stands for encoder-decoder architecture. Autoregressive means that the model was trained with a causal mask. Note that UniLM is trained using a generalized language modeling objective that includes both unidirectional and bidirectional subtasks and cannot be attributed to either autoregressive or non-autoregressive.}
\label{tab:comparison-of-models}
\end{table*}
%
\footnotetext{$\mathrm{GPT_{NEO}}$ is trained on a 800Gb dataset.}

\section{Background}

\subsection{Models}

We use 8 different model families in this study. All of them are based on the transformer architecture and pre-trained on general-domain texts, but this is where the similarities end. We summarize their major differences in Table \ref{tab:comparison-of-models}. In this section, we discuss and highlight the details that distinguish models, from the major ones to the ones that might appear very minor.

\textbf{BERT} \cite{devlin2018bert} is pre-trained on Book Corpus and Wikipedia using a combination of Masked Language Modeling (MLM) and Next Sentence Prediction (NSP). It uses GELU activations \cite{Hendrycks2016BridgingNA} for fully-connected layers. For the first 90\% of the training iterations, the maximum length is 128, but then it is increased to 512.

\textbf{RoBERTa} \cite{liu2019roberta} is the most similar to $\mathrm{BERT}$ in this study; however, it differs from it in many small but important details: the pre-training dataset is considerably larger and includes OpenWebText \cite{Gokaslan2019OpenWeb}, Stories \cite{trinh2018simple}, and CC-News. $\mathrm{RoBERTa}$ does not use Next Sentence Prediction; applies masking dynamically; always trains with 512 max tokens; uses a smaller ADAM $\beta=0.98$; 8 times larger batch size than $\mathrm{BERT}$; and has a larger, byte-level BPE vocabulary (50K instead of 31K).

\textbf{DistilBERT} \cite{sanh2019distilbert} is a distilled version of BERT. It has half the layers of $\mathrm{BERT}$ and is trained using soft targets produced by $\mathrm{BERT}$.

\textbf{ALBERT} \cite{lan2019albert} shares parameters across transformer layers and uses an extra projection between the embedding and the first transformer layer. It replaces NSP with the sentence-order prediction. $\mathrm{ALBERT}$ uses n-gram masking and the LAMB \cite{you2019large} optimizer. The training setup is similar to $\mathrm{BERT}$, but it trains 90\% of the time using the sequence length 512 and randomly reduces it in 10\% of iterations. Parameter sharing allows $\mathrm{ALBERT}$ to achieve performance similar to $\mathrm{BERT}$ with much fewer trainable parameters. The smallest $\mathrm{ALBERT}$ model has 12M trainable parameters and the largest has 235M.

$\textrm{ALBERTv2}$ is a minor modification of $\textrm{ALBERT}$ that was trained without dropout, for twice as many training steps with additional training data \footnote{\href{https://github.com/google-research/albert}{\texttt{github.com/google-research/albert}}}.

\textbf{GPT-2} \cite{radford2019language} is a unidirectional transformer language model trained on the WebText dataset. Unlike other models, it is a Pre-Norm transformer. Similar to $\mathrm{RoBERTa}$, $\mathrm{GPT2}$ has a 50K vocabulary and a byte-level BPE but treats spaces as a separate symbol. It also comes in multiple sizes from 124M parameters up to 2.8B parameters. There exist several popular reimplementations of this model, such as $\textrm{GPT-Neo}$ \cite{gpt-neo}, which generally follow the original paper but differ in dataset \cite{gao2020pile}, model, and training hyperparameters.

\textbf{UniLM} \cite{dong2019unified} utilizes several attention masks to control the access to context for each word token. It uses a multitask objective that is modeled by applying different attention masks. The mix of tasks includes masked language modeling, unidirectional language modeling, and sequence-to-sequence language modeling. Additionally, it employs the NSP objective and is initialized using $\mathrm{BERT}$ model weights. In optimization, it generally follows $\mathrm{BERT}$ but always uses 512 as the maximum sequence length.

\textbf{BART} \cite{lewis2019bart} is an encoder-decoder model that is trained on text infilling and sentence permutation tasks. It is trained on the same dataset as $\mathrm{RoBERTa}$. Compared to $\mathrm{BERT}$, $\mathrm{BART}$ does not use an additional projection when predicting word logits. In optimization, it closely follows $\mathrm{RoBERTa}$, but disables dropout for the final 10\% of training.

\textbf{T5} \cite{raffel2019exploring} is also an encoder-decoder model. It is trained using a text infilling task on the C4 dataset. However, it only generates the text in place of the [MASK] token and not the full input sentence. Architecturally, it is a Pre-Norm model and T5 LayerNorm does not use bias. Output projection weights are tied with the input embedding matrix. It uses 128 relative positional embeddings that are added at every layer. Unlike most of the models in this study, it uses the ReLU activation. The smallest T5 model used in this study has 233M parameters and the largest has 2.8B. We have not evaluated the 11B T5 model due to hardware limitations.

Unlike the original $\mathrm{T5}$, $\mathrm{T5v1.1}$\footnote{\href{https://huggingface.co/google/t5-v1_1-base}{\texttt{huggingface.co/google/t5-v1\_1-base}}} is trained on different data, does not tie logit layer with input embeddings, uses GEGLU activations \cite{shazeer2020glu} and no dropout. It also slightly changes model shapes.






\subsection{oLMpics}
\label{sec:olmpics}
%
The oLMpics benchmark consists of eight tasks that test multiple specific skills, such as a model's ability to draw comparisons, understand negation, and perform simple linguistic composition tasks.
Table \ref{tab:examples} shows examples for every task in oLMpics.

\begingroup
\begin{table*}
\centering
\small
\begin{tabular}{ l | p{8cm} | p{3.5cm}} 
Task Name & Example Question & Choices \\
\midrule
    {Age Comparison} & A 41 year old person age is [MASK] than a 42 year old person. & \underline{younger}, older \\
    {Object Comparison} & The size of a nail is usually [MASK] than the size of a fork. & \underline{smaller}, larger \\
    {Antonym Negation} & It was [MASK] a fracture, it was really a break. & not, \underline{really} \\
    {Taxonomy Conjunction} & A ferry and a biplane are both a type of [MASK]. & airplane, \underline{craft}, boat \\
    Property Conjunction & What is related to vertical and is related to honest? & \underline{straight}, trustworthy, steep \\
    Encyclopedic Composition & When did the band where Alan Vega played first form? & \underline{1970}, 1968, 1969 \\
    Hypernym Conjunction &  A basset and a tamarin are both a type of [MASK] & primate, dog, \underline{mammal}\\
    {Multi-hop Composition} & When comparing a 21 year old, 15 year old, and 19 year old, the [MASK] is oldest. & third, \underline{first}, second \\
\bottomrule
\end{tabular}
\caption{Examples of oLMpics questions, with the \underline{correct answer} underlined.}
\label{tab:examples}
\end{table*}
\endgroup

\paragraph{Zero-Shot vs. Multi-Shot}
A major advantage of the oLMpics tasks is that zero-shot evaluation can be performed for most tasks due to the task format. Zero-shot evaluation eliminates the ambiguity of whether a model's knowledge is stored in its pre-trained representations or learned during fine-tuning. However, a model may possess the necessary information but fail during zero-shot evaluation due to the wording of the task. Therefore, multi-shot evaluation can also be informative, allowing the model to adapt to the input format and possibly learn task-specific features.
OLMpics tasks include training sets specifically for this reason, in order to separate the impact of fine-tuning from pre-training.
%

\paragraph{MC-MLM vs. MC-QA}
The oLMpics tasks are framed in one of two ways: MC-MLM (Multiple Choice-Masked Language Modeling) and MC-QA (Multiple Choice-Question Answering).
MC-MLM tasks are formulated as a masked language modeling task \cite{devlin2018bert}, where the model needs to predict the word replaced by the MASK token.
An example of an \textit{Age Comparison} sentence is ``A 41 year old is [MASK] a 42 year old.''
A model's prediction is determined by the probabilities assigned to the [MASK] token, with ``younger'' being selected if its probability is higher than ``older,'' and ``older'' otherwise.

MC-MLM restricts the possible answers to single tokens. Tasks with longer answers require MC-QA. In this method, a new feedforward network maps the [CLS] token embedding to a single logit. For prediction, answer choices are individually concatenated to the original question, forming a new sentence for each choice.
This set of sentences is input into the model, and the choice corresponding to the sentence with the largest logit is selected.
While the MC-QA method allows for longer choices, the added feedforward network must be trained; therefore, zero-shot evaluation is not possible.

\paragraph{Extending Beyond MLM}
The oLMpics MC-MLM method relies on the model giving probabilities of individual words in a bidirectional context.
However, models like $\mathrm{GPT2}$ do not have access to the future context, which makes it impossible to directly predict the token in an example like ``A 41 year old is [MASK] than 42 year old.''
For these models, we sum the log-probabilities of individual words to find the probability of the whole sentence. We do this for every possible answer, e.g., a sequence with ``younger'' instead of [MASK] and ``older''. Then, we select the one with the highest total probability.

Extending BART and T5 is more straightforward because their objectives and architecture are very flexible.
For both of these models, we use the original oLMpics input format. T5 has multiple [MASK]-tokens and we always use \texttt{<extra\_id\_0>} token in our evaluation.
The biggest difference is that BART produces the full sentence and we need to extract the probabilities for the masked words and T5 produces only the tokens in the place of [MASK].

\subsection{Psycholinguistic Data}

Similar to oLMpics, the datasets used by  \citet{ettinger2020bert} are framed as ``fill in the blank'' tasks. Unlike oLMpics, the model always needs to predict only the last word, so both bidirectional and unidirectional models can be evaluated on these tasks directly.
The biggest distinction of this dataset is its source.
The datasets CPRAG-102 \cite{federmeier1999rose}, ROLE-88 \cite{chow2016bag}, and NEG-136 \cite{fischler1983brain} come from the psycholinguistics and neuroscience studies and were originally evaluated on humans.





CPRAG-102 targets commonsense and pragmatic inference e.g. \textit{Justin put a second house on Park Place. He and his sister often spent hours playing \_\_}, Target: \textit{monopoly}, other labels: \textit{chess, baseball}. ROLE-88 aims at evaluating event knowledge and semantic roles.

NEG-136 tests how well models understand the meaning of negation and consists of two subsets: simple (SIMP) and natural (NAT).
For example, SIMP: \textit{Salmon is a \underline{fish/dog}} versus \textit{Salmon is not a \underline{fish/dog.}} NAT: \textit{Rockets and missiles are very \underline{fast/slow}} versus \textit{Rockets and missiles aren't very \underline{fast/slow}}.
Evaluation of this dataset is performed in two ways: affirmative statements and negative statements. For affirmative ones, the model needs to complete a sentence like \textit{A robin is a} with the expected answer \textit{bird}. For negative, \textit{A robin is not a} should not be completed with a \textit{bird}. \cite{ettinger2020bert} finds that this type of error is very common in $\mathrm{BERT}$, which suggests that the model cannot handle negation correctly.

\citet{ettinger2020bert} tests $\mathrm{BERT}$ models in two ways: using a pre-defined set of answers, similar to oLMpics MC-MLM, or computing top-k accuracy from the whole model vocabulary. We adopt the same approach in this study.

\section{Experiments}

We evaluate eight models families on the oLMpics (29 models in total) and six families on psycholinguistic data (17 models). This extends the \citet{talmor2019olmpics} results with six new model families and \citet{ettinger2020bert} with four.

\begingroup
\begin{table*}
\centering
\small
\begin{tabular}{l | c c c c} 
\toprule
Input sequence example & $\mathrm{GPT2}_{B}$ & $\mathrm{GPT2}_{M}$ & $\mathrm{GPT2}_{L}$ & $\mathrm{GPT2}_{XL}$\\
\midrule
     (oLMpics) It was \textcolor{red}{really}/\textcolor{teal}{not} sane, it was really insane & 53.3 & 52.8 & \textbf{59.0} & 60.6 \\
\midrule
     It was really insane. Was it sane ? \textcolor{red}{yes}/\textcolor{teal}{no}& 51.6 & \textbf{58.2} & 55.6 & \textbf{61.4} \\
     It was really insane. Was it really sane ? \textcolor{red}{yes}/\textcolor{teal}{no} & 50.2 & 54 & 50.2 & 54.4 \\
     It was sane entails it was really insane ? \textcolor{red}{yes}/\textcolor{teal}{no} & 49.8 & 50.2 & 50 & 50.6 \\
     Sentence 1: It was sane. Sentence 2:  It was really insane. &  \textbf{59.6} & 50.2 & 46.8 & 48.4 \\
     Is Sentence 1   synonym of Sentence 2? \textcolor{red}{yes}/\textcolor{teal}{no} \\
\bottomrule
\end{tabular}
\caption{Prompts for the Antonym Negation task. Random baseline accuracy is 50\%. The original oLMpics prompt is the prompt used in Table \ref{tab:mc-mlm-zero-shot}. $\mathrm{GPT2}_{B}$ is the base-sized model, $\mathrm{GPT2}_{M}$ is medium, and $\mathrm{GPT2}_{L}$ is large. Text highlighted in red/green are correct/wrong labels.}
\label{tab:prompts-antonym-negation}
\end{table*}
\endgroup

\begin{table*}[h!]
\centering
\small
\begin{tabular}{ l | c c c c c c c c}
\toprule
& \thead{Age \\  Comp.} & \thead{Always \\  Never} & \thead{Object \\ Comp.} & \thead{Antonym \\ Negation} & \thead{Taxonomy \\ Conj.} & \thead{Multi-hop \\ Comp.} \\
\midrule
        Majority & 50.6 & 36.1 & 50.6 & 50.2 & 34.0 & 34.0 \\
        \midrule
        $\mathrm{BERT}_\mathrm{base}$ & 49.4 & 13.3 & 55.4 & 53.8 & 46.7 & 33.2 \\ 
        $\mathrm{BERT}_\mathrm{large}$ & 50.6 & 22.5 & 52.4 & 51.0 & \textbf{53.9} & 33.8 \\ 
        $\mathrm{BERT}_\mathrm{large}$ WWM & 76.6 & 10.7 & 55.6 & 57.2 & 46.2 & 33.8 \\ 
        $\mathrm{RoBERTa}_\mathrm{large}$ & 98.6 & 13.5 & 87.4 & \textbf{74.4} & 45.4 & 28.0 \\ 
        \midrule
        $\mathrm{DistilBERT}_\mathrm{base}$ & 49.4 & 15.0 & 50.8 & 50.8 & 46.9 & 33.4 \\ 
        $\mathrm{AlBERT}_\mathrm{base}$ & 47.0 & 23.2 & 50.6 & 52.6 & - & 34.0 \\ 
        $\mathrm{AlBERT}_\mathrm{large}$ & 52.8 & 30.7 & 49.2 & 50.2 & - & 34.0 \\ 
        $\mathrm{AlBERT}_\mathrm{xlarge}$ & 39.8 & 26.1 & 50.4 & 44.6 & - & 32.2 \\ 
        $\mathrm{AlBERT}_\mathrm{xxlarge}$ & 95.4 & 22.9 & 61.0 & 66.4 & - & 34.0 \\ 
        $\mathrm{AlBERTv2}_\mathrm{base}$ & 50.6 & 21.4 & 49.4 & 54.2 & - & 14.0 \\ 
        $\mathrm{AlBERTv2}_\mathrm{large}$ & 51.4 & 31.7 & 50.6 & 55.2 & - & 34.0 \\ 
        $\mathrm{AlBERTv2}_\mathrm{xlarge}$ & 46.2 & \textbf{37.9} & 50.6 & 62.4 & - & 32.4 \\ 
        $\mathrm{AlBERTv2}_\mathrm{xxlarge}$ & 93.8 & 23.9 & 78.8 & 64.8 & - & 34.0 \\ 
        $\mathrm{BART}_\mathrm{large}$ & 86.0 & 14.3 & 50.8 & 53.8 & 42.6 & 33.8 \\ 
        $\mathrm{T5}_\mathrm{small}$ & 49.4 & 16.1 & 48.2 & 47.0 & 49.3 & 33.8 \\ 
        $\mathrm{T5}_\mathrm{base}$ & 49.4 & 10.7 & 59.0 & 53.4 & 46.6 & 33.6 \\ 
        $\mathrm{T5}_\mathrm{large}$ & 94.0 & 25.7 & 79.8 & 59.2 & 42.2 & 33.8 \\ 
        $\mathrm{T5}_\mathrm{xl}$ & \textbf{100.0} & 20.4 & \textbf{90.0} & 68.4 & 41.2 & 34.4 \\ 
        $\mathrm{T5v1.1}_\mathrm{small}$ & 49.4 & 34.3 & 50.6 & 51.4 & 48.2 & \textbf{37.8} \\ 
        $\mathrm{T5v1.1}_\mathrm{base}$ & 50.6 & 11.8 & 56.0 & 45.0 & 49.9 & 37.6 \\ 
        $\mathrm{T5v1.1}_\mathrm{large}$ & 49.6 & 15.7 & 50.6 & 47.0 & 41.7 & 33.8 \\ 
        $\mathrm{T5v1.1}_\mathrm{xl}$ & 49.4 & 23.9 & 49.4 & 54.2 & \textbf{53.9} & 33.8 \\ 
        $\mathrm{UniLM}_\mathrm{base}$ & 47.9 & 15.5 & 47.8 & 43.5 & 45.1 & 34.9 \\ 
        $\mathrm{UniLM}_\mathrm{large}$ & 47.9 & 19.2 & 61.1 & 50.8 & 50.2 & 33.1 \\
        $\mathrm{GPT2}_\mathrm{base-0.1B}$ & 47.6 & 9.0 & 50.3 & 53.3 &  49.1 & 32.6 \\
        $\mathrm{GPT2}_\mathrm{medium-0.3B}$ & 50.1 & 31.3 & 50.3 & 52.8 & 51.9 & 34.0 \\ 
        $\mathrm{GPT2}_\mathrm{large-0.8B}$ & 69.6 & 26.0 & 50.5 & 59.0 &  46.9 & 34.0 \\ 
        $\mathrm{GPT}_\mathrm{NEO-1.3B}$ & 58.6 & 29.0 & 52.1 & 65.2 & 50.6 & 33.3 \\
        $\mathrm{GPT2}_\mathrm{xl-1.5B}$ & 51.9 & 26.6 & 52.6 & 60.6 & 45.8 & 34.0 \\
        \bottomrule
\end{tabular}
\caption{Zero-shot oLMpics evaluation on MC-MLM tasks. ``Majority'' here is the accuracy when predicting the most frequent class. The first 4 models are our reproduction of the original oLMpics results.
The best result on each task is highlighted in bold.
We do not evaluate $\mathrm{ALBERT}$ on Taxonomy Conjunction because its vocabulary does not contain classes as single tokens. A version of this table with confidence intervals can be found in Table \ref{tab:mc-mlm-zero-shot-conf-intervals} in the Appendix.}.
\label{tab:mc-mlm-zero-shot}
\end{table*}

\subsection{Language models are not universal multitask learners}
\label{sec:lanuage_models_not_universal}

\paragraph{Zero-shot evaluation}
It has been shown that language models can implicitly learn downstream tasks \cite{radford2019language,brown2020language_gpt3}.
However, it is still not obvious what tasks are learnable in this manner without explicit supervision. In our study, similar to \citet{talmor2019olmpics}, we find that none of the models can solve Multi-Hop Composition or Always-Never tasks substantially better than a majority baseline (see Table \ref{tab:mc-mlm-zero-shot}).

This holds true not only for masked language models but also for unidirectional language models such as $\mathrm{GPT2}$ and text-infilling models such as $\mathrm{T5}$ or $\mathrm{BART}$.
Only $\mathrm{small}$ and $\mathrm{base}$ versions of $\mathrm{T5v1.1}$ outperform the majority baseline on \textit{Multi-Hop Composition} by a small margin.


\paragraph{Multi-shot evaluation}

Not surprisingly, fine-tuning models on oLMpics improves the scores across the board.
This is true even for the tasks on which zero-shot performance is extremely poor.
For example, while all models fail on \textit{Multi-hop Composition} during zero-shot evaluation, most models can reach perfect or near-perfect accuracy on this task after fine-tuning.    
However, \textit{Always-Never} and \textit{Taxonomy Conjunction} remain challenging for all models.
For the full multi-shot evaluation, see Table \ref{tab:mc-qa} in the Appendix.



\subsection{Bigger does not mean better}
To check how the size of a model affects the performance, we evaluated different versions of $\mathrm{GPT2}$, $\mathrm{T5}$, and $\mathrm{ALBERT}$ models on the oLMpics tasks
ranging from 14M (smallest $\mathrm{ALBERT}$) to 2.8B (largest $\mathrm{T5}$) parameters.
All of the models perform near-random on 3 out of the 6 tasks, suggesting that Multi-Hop Composition, Antonym Negation, and Always-Never are hard to learn via the (masked) language modeling objective.
On the rest of the tasks, we observe no clear improvement trend for $\mathrm{GPT}$ models based on the model size.
In most of the tasks, $\mathrm{GPT_{large}}$ either performs on par or has higher accuracy than $\mathrm{GPT_{xl}}$ while being twice as small.

We also compute Spearman correlation between model accuracy and model size for $\mathrm{GPT2}$, $\mathrm{ALBERT}$, and $\mathrm{T5}$ models.\footnote{Note that sample size for each test is $\leq$ 4, so these results should be taken as anecdotal.}
For all $\mathrm{GPT2}$ and $\mathrm{ALBERT}$ (v1 and v2) tests, the p-value is $\gg 0.05$, suggesting that there is no rank-correlation between model size and task performance.
However, in the case of T5 models, there is a strong (1.0) and significant correlation (p-value $\sim 10^{-6}$) for all tasks except \textit{Always-Never}. We account for multiple hypothesis testing using Bonferroni’s method.
For \textit{Taxonomy Conjunction}, the correlation is negative.

\subsection{Model properties are not predictive of model performance}
\label{sec:model_properties_are_not_predictive}

Contrary to the common knowledge, with rare exceptions (Section \ref{sec:lanuage_models_not_universal}), we do not observe that parameter count, dataset size, model architecture or directionality are predictive of model performance on zero-shot oLMpics (Table \ref{tab:mc-mlm-zero-shot}).


$\mathrm{RoBERTa_{large}}$ usually performs amongst the best models, while having a very similar architecture and objective to $ \mathrm{BERT_{large}}$.
Reasonable explanations would be the dataset size, but this does not align with the $\mathrm{BART_{large}}$ results.
Encoder-decoder architecture does seem not to be indicative of the performance either, as $ \mathrm{T5_{large}}$ and $ \mathrm{BART_{large}}$ have vastly different results.

Psycholinguistic datasets (Table \ref{tab:ettinger-word_prediction}) demonstrate similar behaviour. $\mathrm{RoBERTa_{large}}$ is generally the stongest model followed by $\mathrm{T5_{xl}}$. We would like to note that these datasets have less than 100 examples and their statistical power \cite{card-etal-2020-little} is very small.



Our intuitions about the relative suitability of different model classes are based on their performance on standard benchmarks  \cite{wang2018glue,wang2019superglue} and existing investigations of scaling laws \cite{radford2019language,kaplan2020scaling}. In contrast to this received wisdom, our experiments suggest that this does not in fact lead to better performance on specific linguistic skills.


\subsection{RoBERTa is sensitive to negation}

\citet{ettinger2020bert} observed that $\mathrm{BERT}$ is not sensitive to negation in non-natural (SIMP) or less-natural cases.
In our experiments (Table \ref{tab:ettinger-sensitivity_prediction}), we find that the only model with zero accuracy outside of $\mathrm{BERT}$ is a distilled version of $\mathrm{BERT}$ itself.
Multiple models achieve non-zero accuracy on NEG-SIMP (neg), but the numbers might be misleading.
For example, while $\mathrm{ALBERTv1_{xlarge}}$ has 27.8\% accuracy on NEG-SIMP (neg), this accuracy is mainly caused by mistakes in language modeling while still being insensitive to negation (e.g., it predicts \textit{vegetable} for both \textit{An ant is a} and \textit{An ant is not a}). Specifically, $\mathrm{ALBERTv1_{xlarge}}$ only changes its predictions in 5.5\% cases.

However, unlike other models, $\mathrm{RoBERTa_{large}}$ actually changes its predictions in 33\% cases, suggesting that sensitivity to negation might be possible to learn via masked language modeling.

\subsection{Models make plausible mistakes}

One drawback of datasets from \citet{ettinger2020bert} that we have noticed was the ambiguity of answers.
For example, many models predict words like ``this'', ``that'', ``it'' as the next word for \textit{“Checkmate,” Rosaline announced with glee. She was getting to be really good at [MASK]} instead of the word ``chess''.
In fact, for $\mathrm{T5}_\mathrm{xl}$ predictions, we found that 79.4\% of predictions are semantically and grammatically plausible, while this model has only achieved 58.8\% top-5 accuracy on the CPRAG-126 dataset (Table \ref{tab:ettinger-word_prediction}).

Another example would be \textit{I’m an animal like Eeyore!” the child exclaimed. His mother wondered why he was pretending to be a [MASK]}. CPRAG expects the answer ``donkey'', which assumes that the reader (or model) is familiar with the English names of Winnie-the-Pooh book characters.\footnote{Only one of the authors of this paper was able to continue this sentence correctly}

\begingroup
\begin{table*}
\centering
\small
\begin{tabular}{ l | c c c c}
\toprule
& \thead{CPRAG-126} & \thead{ROLE-88} & \thead{NEG-136\\SIMP(Aff)} & \thead{NEG-136\\NAT(Aff)} \\
\midrule
        $\mathrm{BERT}_\mathrm{base}$ & 52.9 & 27.3 & \textbf{100} & 43.8 \\ 
        $\mathrm{BERT}_\mathrm{large}$ & 52.9 & 37.5 & \textbf{100} & 31.3 \\
        \midrule
        $\mathrm{RoBERTa}_\mathrm{base}$ & 70.1 & 46.6 & 94.4 & 56.3 \\ 
        $\mathrm{RoBERTa}_\mathrm{large}$ & 82.4 & \textbf{55.7} & 94.4 & 50 \\ 
        $\mathrm{DistilBERT}_\mathrm{base}$ & 55.9 & 28.4 & 94.4 & 43.8 \\ 
        $\mathrm{AlBERTv1}_\mathrm{base}$ & 17.6 & 17.1 & 72.2 & 25.0 \\
        $\mathrm{AlBERTv1}_\mathrm{large}$ & 35.3 & 26.1 & 83.3 & 25 \\
        $\mathrm{AlBERTv1}_\mathrm{xlarge}$ & 41.2 & 34.1 & 55.5 & 18.8 \\
        $\mathrm{AlBERTv1}_\mathrm{xxlarge}$ & 82.4 & 53.4 & 72.2 & 50 \\
        $\mathrm{AlBERTv2}_\mathrm{base}$ &  41.4 & 26.1 & 33.3 & 31.1 \\ 
        $\mathrm{AlBERTv2}_\mathrm{large}$ & 47.1 & 29.5 & 83.3 & 37.5 \\
        $\mathrm{AlBERTv2}_\mathrm{xlarge}$ & 61.8 & 37.5 & 94.4 & 25 \\
        $\mathrm{AlBERTv2}_\mathrm{xxlarge}$ & \textbf{85.3} & 50 & \textbf{100} & 37.5 \\
        $\mathrm{T5}_\mathrm{small}$ & 20.6 & 9.1 & 44.4 & 18.8 \\
        $\mathrm{T5}_\mathrm{base}$ & 41.1 & 27.3 & 88.9 & 31.3 \\
        $\mathrm{T5}_\mathrm{large}$ & 50.0 & 36.4 & 94.4 & 43.8 \\
        $\mathrm{T5}_\mathrm{xl}$ & 58.8 & 44.3 & 83.3 & \textbf{62.5} \\
        \bottomrule
\end{tabular}
\caption{Zero-shot top-5 word prediction accuracy. Top-5 is selected over the whole model vocabulary. 
The best result on each task is highlighted in bold. SIMP stands for simple, NAT stands for natural. Both negation tasks are evaluated in the affirmative form. The first 2 models are our reproduction of the original results. 
}
\label{tab:ettinger-word_prediction}
\end{table*}
\endgroup

\begingroup
\begin{table*}
\centering
\small
\begin{tabular}{ l | c c c c c c c c}
\toprule
& \thead{CPRAG} & \thead{ROLE} & \thead{NEG\\SIMP\\(Aff)} & \thead{NEG\\SIMP\\(Neg)} & \thead{NEG\\NAT\\(Aff)} & \thead{NEG\\NAT\\(Neg)} & \thead{NEG\\LNAT\\(Aff)} & \thead{NEG\\LNAT\\(Neg)}
\\
\midrule
        $\mathrm{BERT}_\mathrm{base}$ & 73.5 & 75.0 & \textbf{100.0} & 0.0 & 62.5 & 87.5 & \textbf{75.0} & 0.0 \\
        $\mathrm{BERT}_\mathrm{large}$ & \textbf{79.4} & \textbf{86.4} & \textbf{100.0} & 0.0 & \textbf{75.0} & \textbf{100} & \textbf{75.0} & 0.0 \\
        \midrule
        $\mathrm{RoBERTa}_\mathrm{base}$ & 23.5 & 50.0 & 66.7 & 33.3 & 25 & 75.0 & \textbf{75.0} & 12.5 \\ 
        $\mathrm{RoBERTa}_\mathrm{large}$ & 29.4 & 56.8 & 66.7 & 33.3 & 37.5 & 75.0 & \textbf{75.0} & 12.5 \\ 
        $\mathrm{DistilBERT}_\mathrm{base}$ & 70.6 & 72.8 & \textbf{100.0} & 0.0 & \textbf{75.0} & 43.8 & 43.8 & 18.9\\ 
        $\mathrm{AlBERTv1}_\mathrm{base}$ & 11.8 & 40.1 & 77.8 & 16.4 & 25.0 & 25.0 & \textbf{75.0} & 37.5\\
        $\mathrm{AlBERTv1}_\mathrm{large}$ & 23.5 & 43.2 & 88.8 & 16.7 & 25 & 50 & \textbf{75.0} & 12.5 \\
        $\mathrm{AlBERTv1}_\mathrm{xlarge}$ & 17.6 & 52.3 & 61.1 & 27.8 & 25.0 & 50.0 & \textbf{75.0} & 12.5\\
        $\mathrm{AlBERTv1}_\mathrm{xxlarge}$ & 32.3 & 56.8 & 88.9 & 16.7 & 25.0 & 62.5 & \textbf{75.0} & 12.5 \\
        $\mathrm{AlBERTv2}_\mathrm{base}$ & 20.1 & 56.8 & 72.2 & 22.2 & 25.0 & 50.0 & \textbf{75.0} & 25.0  \\ 
        $\mathrm{AlBERTv2}_\mathrm{large}$ &29.4 & 54.5 & 83.3 & 11.1 & 25.0 & 62.5 & \textbf{75.0} & 12.5 \\
        $\mathrm{AlBERTv2}_\mathrm{xlarge}$ & 20.6 & 61.4 & 83.3 & 16.7 & 25.0 & 62.5 & \textbf{75.0} & 25.0 \\
        $\mathrm{AlBERTv2}_\mathrm{xxlarge}$ & 32.4 & 54.5 & 83.3 & 16.7 & 37.5 & 62.5 & \textbf{75.0} & 12.5 \\
        $\mathrm{T5}_\mathrm{small}$ & 5.9 & 45.5 & 55.6 & 33.3 & 50.0 & 25.0 & 37.5 & \textbf{62.5} \\
        $\mathrm{T5}_\mathrm{base}$ & 14.7 & 70.5 & 61.1 & 27.8 & 50.0 & 12.5 & 37.5 & 37.5 \\
        $\mathrm{T5}_\mathrm{large}$ & 17.6 & 54.5 & 72.2 & 16.7 & 62.5 & 37.5 & 37.5 & 50.0 \\
        $\mathrm{T5}_\mathrm{xl}$ & 14.7 & 63.6 & 66.7 & 27.8 & 62.5 & 50.0 & 37.5 & 50.0 \\
        
        $\mathrm{GPT2}_\mathrm{base}$ & 11.8 & 34.1 & 66.7 & \textbf{38.9} & \textbf{75.0} & 25.0 & 37.5 & 37.5 \\
        $\mathrm{GPT2}_\mathrm{medium}$ & 17.6 & 36.4 & 61.1 & 22.2 & 50.0 & 50.0 & 50.0 & \textbf{62.5} \\
        $\mathrm{GPT2}_\mathrm{large}$ & 29.4 & 45.5 & 77.8 & 16.7 & 62.5 & 50.0 & 37.5 & 50.0 \\
        $\mathrm{GPT}_\mathrm{neo}$ & 20.6 & 45.5 & 77.8 & 33.3 & \textbf{75.0} & 37.5 & 62.5 & 25.0 \\
        $\mathrm{GPT2}_\mathrm{xl}$ & 17.6 & 50.0 & 61.1 & 33.3 & 62.5 & 75.0 & 62.5 & 37.5 \\
        
        \bottomrule
\end{tabular}
\caption{Zero-shot accuracy on tasks from \citet{ettinger2020bert}. Accuracy is measured as the percentage of instances for which the model assigns a higher probability to the good completion than to the bad completions (pre-defined). The best result on each task is highlighted in bold. SIMP stands for simple, NAT for natural, LNAT for less natural as defined in the original paper. The first 2 models are our reproduction of the original results.
}
\label{tab:ettinger-sensitivity_prediction}
\end{table*}
\endgroup



\subsection{Antonym Negation: Impact of prompt variation}

While there is clear evidence that models pre-trained with the MLM objective have trouble with negation \cite{ettinger2020bert}, no such evidence has been available for models trained autoregressively. At the same time, a number of studies have shown that autoregressive models can be significantly improved with prompting.
Our question is whether we can make a language model (GPT-2) understand negation via an alternative wording of the task (prompt engineering).

We tested four different prompts for the Antonym Negation task.
Table \ref{tab:prompts-antonym-negation} shows the patterns and the corresponding accuracies of $\mathrm{GPT}$ models.
All experiments use ``yes''/``no'' verbalizers.
While some prompts improve the oLMpics prompt results (up to +6\%), this improvement is not consistent across models showing that even very similar models are sensitive to prompt variation in different ways.

Additionally, prompt \#4 (Table \ref{tab:prompts-antonym-negation}) improves the smallest model, $\mathrm{GPT2_{base}}$, so significantly that it outperforms the largest model by approximately 10\%, demonstrating once again that parameter count is not a reliable predictor of the model performance.

Table \ref{tab:prompts-antonym-negation} shows the GPT2 results for these four patterns. The reformatting helped GPT2 model perform better. When compared with Table~\ref{tab:mc-mlm-zero-shot}, there was an improvement of around 9\%, 8\% and 5\% in GPT2, GPT-medium and GPT2-large respectively.
However none of these models surpasses or reach close to the performance of RoBERTa.

\begin{figure}
    \centering
    \includegraphics[width=0.48\textwidth]{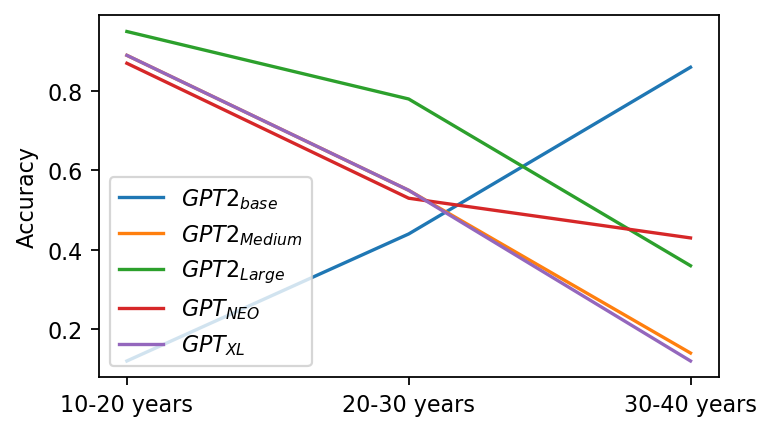}
    \caption{Evaluation of $\mathrm{GPT2}$ variants on Age Comparison task for different age groups. }
    \label{fig:ge-group-comparison-gpt2-plot}
\end{figure}

\subsection{Age Comparison: Accuracy varies by age group}
For one oLMpics task, Age Comparison, we observe that models do not perform equally well on all age ranges, similar to the findings of \citet{talmor2019olmpics}. Figure \ref{fig:ge-group-comparison-gpt2-plot} shows that with the exception of $\mathrm{GPT2}_\mathrm{base}$, all $\mathrm{GPT}2$ variants perform well on 10-20 year olds and poorly on the 30-40 age group, with a significant drop in performance from 80\% to 20\%.
Generally, $\mathrm{GPT2}$ seems to predict younger ages more accurately.
However, the smallest model, $\mathrm{GPT2}_\mathrm{base}$, exhibits a different trend than other models as age increases.

This is interesting, because all $\mathrm{GPT2}$ models, except $\mathrm{GPT2_{NEO}}$, were trained on the same dataset and only vary in size.

\subsection{Model performance is highly sensitive to punctuation}

\begin{figure}
    \centering
    \includegraphics[width=0.48\textwidth]{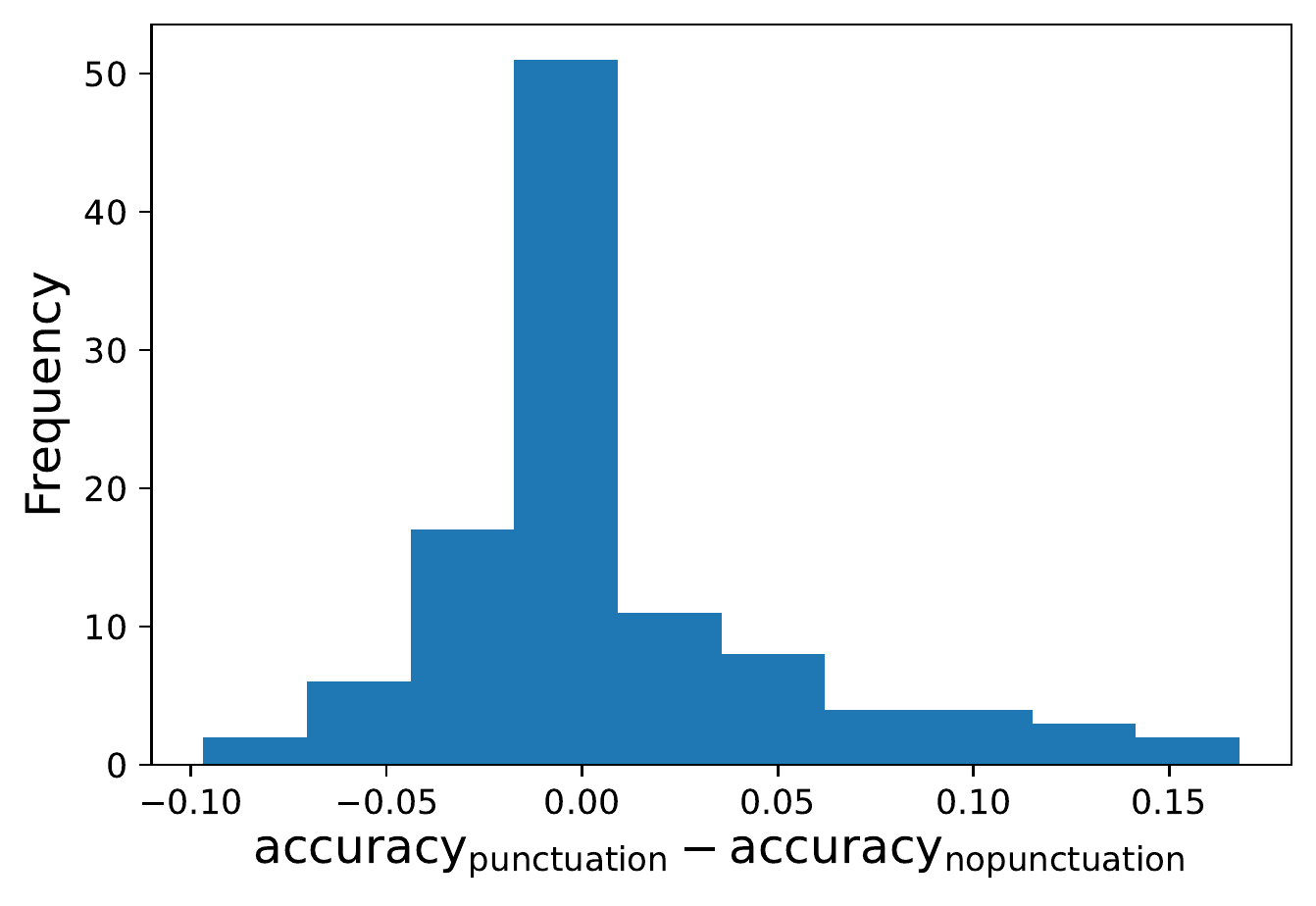}
    \caption{Effect of having a full stop symbol at the end of examples on accuracy on oLMpics datasets.}
    \label{fig:punct_hist}
\end{figure}

We find that model performance can change significantly on both oLMpics and psycholinguistic datasets if we add a period to the end of the sequence. For example, $\mathrm{BERT}$ and $\mathrm{DistilBERT}$ achieve an accuracy of 3\% without a period on CPRAG as compared to 52.9\% when a '.` is appended.
We observe a similar trend on the ROLE and NEG datasets and for other models including $\mathrm{RoBERTa}$, where the accuracy on CPRAG jumped from 47.1\% to 70.1\%.
For oLMpics, the change of performance is less dramatic, but still noticeable. We observe that in 6\% of cases (across all models and all tasks), model performance changes by more than 10 absolute percentage points if a full stop is added to the end of sentence. Figure \ref{fig:punct_hist} shows the histogram of accuracy changes for oLMpics tasks.

\section{Conclusion}
In this work, we apply a large and diverse set of models to oLMpics and psycholinguistic tasks. 
The variety of models allows us to investigate the performance of different architectures and pre-training methods on a variety of linguistic tasks.
%

Contrary to received wisdom, we find that parameter count within a given model family does not correlate with model performance on these tasks.
We find that none of the models, even the 2.8B-sized ones, can resolve \textit{Multi-Hop Composition} and \textit{Always-Never} tasks in a zero-shot manner, suggesting that the existing pre-training methods cannot learn such tasks.
Finally, we find that different models excel in different symbolic reasoning tasks, suggesting that slight differences related to optimization or masking strategy might be more important than the pre-training approach, dataset size, or architecture.







\bibliography{anthology,custom}
\bibliographystyle{acl_natbib}

\clearpage
\appendix

\section{Additional Tables}
The next pages present additional results, including the version of Table \ref{tab:mc-mlm-zero-shot} with confidence intervals (Table \ref{tab:mc-mlm-zero-shot-conf-intervals}), oLMpics MC-QA results (Table \ref{tab:mc-qa}), T5 zero-shot Encyclopedic Composition and Property Conjunction (Table \ref{tab:t5-qa}), and T5 evaluated on psycholinguistic datasets when removing stop-words from the model output vocabulary (Table \ref{tab:t5-stop-word}).

\begingroup
\begin{table*}[h!]
\centering
\small
\begin{tabular}{ l | c c c c c c c c}
\toprule
& \thead{Age \\  Comp.} & \thead{Always \\  Never} & \thead{Object \\ Comp.} & \thead{Ant. \\ Neg.} & \thead{Tax. \\ Conj.} & \thead{Multi-hop \\ Compos.} & \thead{Encyc. \\Conj.} & \thead{Prop. \\Conj.} \\
\midrule
        Majority & 50.6 & 20.0 & 50.0 & 50.0 & 34.0 & 34.0 & 34.0 & 34.0 \\
        \midrule
        $\mathrm{BERT}_\mathrm{base}$ & 86.8 & 59.3 & 86.6 & 92.0 & 57.4 & 86.0 & 56.1 & \textbf{62.6} \\
        $\mathrm{BERT}_\mathrm{large}$ & 98.8 & 58.9 & 90.4 & 94.8 & \textbf{60.8} & 99.0 & 57.1 & 58.3 \\
        $\mathrm{BERT}_\mathrm{large}$ WWM & \textbf{100.0} & 58.9 & 85.0 & 95.0 & 58.8 & 97.6 & 56.4 & 60.1 \\
        $\mathrm{RoBERTa}_\mathrm{large}$ & 100 & \textbf{60.4} & 87.2 & \textbf{96.2} & 59.9 & \textbf{100.0} & 55.5 & 55.5 \\
        $\mathrm{DistilBERT}_\mathrm{base}$ & 66.2 & 60.0 & 84.2 & 90.6 & 55.9 & 59.4 & 53.9 & 56.2 \\
        \midrule
        $\mathrm{AlBERT}_\mathrm{large}$ & 91.6 & 59.3 & 66.4 & 90.4 & - & 80.0 & \textbf{57.2} & 60.2 \\
        $\mathrm{BART}_\mathrm{large}$ & \textbf{100.0} & 36.1 & 85.6 & 95.0 & 59.8 & \textbf{100.0} & - & -\\
        $\mathrm{T5}_\mathrm{base}$ & 77.6 & 55.7 & 91.4 & 94.4 & - & 64.8 & - & - \\
        $\mathrm{T5}_\mathrm{large}$ & \textbf{100.0} & 57.9 & \textbf{93.2} & 96.0 & - & \textbf{100.0} & - & - \\
\bottomrule
\end{tabular}
\caption{Multi-shot oLMpics evaluation on MC-MLM and MC-QA tasks. ``Majority'' here is the accuracy when predicting the most frequent class.}
\label{tab:mc-qa}
\end{table*}
\endgroup

\begingroup
\begin{table*}
\centering
\small
\begin{tabular}{ l | c c}
\toprule
& Encyc. Conj. & Prop. Conj. \\
\midrule
        $\mathrm{T5}_\mathrm{small}$ & 29.0 & 38.72 \\ 
        $\mathrm{T5}_\mathrm{base}$ & 31.4 & 36.2 \\ 
        $\mathrm{T5}_\mathrm{large}$ & 31.6 & 34.6 \\ 
        $\mathrm{T5}_\mathrm{xl}$ & 31.2 & 38.5 \\ 
        $\mathrm{T5v1.1}_\mathrm{small}$ & 33.4 & 38.1 \\ 
        $\mathrm{T5v1.1}_\mathrm{base}$ & 31.6 & 40.0 \\ 
        $\mathrm{T5v1.1}_\mathrm{large}$ & 31.4 & \textbf{40.1} \\ 
        $\mathrm{T5v1.1}_\mathrm{xl}$ & \textbf{33.4} & 37.1\\ 
\bottomrule
\end{tabular}
\caption{Zero-shot T5 results on MC-QA tasks. As for T5 can generate multiple tokens in place of a single mask, we evaluate in using similar to MC-MLM. In order to get the probability of the answer, we multiply the probabilities for every answer word.}
\label{tab:t5-qa}
\end{table*}
\endgroup

\begingroup
\begin{table*}
\centering
\small
\begin{tabular}{ l | c c c c}
\toprule
& \thead{CPRAG-126} & \thead{ROLE-88} & \thead{NEG-136\\SIMP(Aff)} & \thead{NEG-136\\NAT(Aff)} \\
\midrule
        $\mathrm{T5}_\mathrm{small}$ & 20.6 & 9.1 & 44.4 & 18.8 \\
        $\mathrm{T5}_\mathrm{base}$ & 38.2 & 22.7 & 88.9 & 31.3 \\
        $\mathrm{T5}_\mathrm{large}$ & 50.0 & 36.4 & \textbf{94.4} & 43.8 \\
        $\mathrm{T5}_\mathrm{xl}$ & 55.9 & 44.3 & 83.3 & \textbf{62.5} \\
\midrule
        $\mathrm{T5}_\mathrm{small}$ Filtered & 20.6 & 15.9 & 55.6 & 25.0 \\ 
        $\mathrm{T5}_\mathrm{base}$ Filtered & 42.2 & 34.1 & 88.9 & 37.5 \\ 
        $\mathrm{T5}_\mathrm{large}$ Filtered & 52.9 & 38.6 & \textbf{94.4} & 43.8 \\ 
        $\mathrm{T5}_\mathrm{xl}$ Filtered & \textbf{58.8} & \textbf{51.1} & 88.9 & \textbf{62.5} \\ 
\bottomrule
\end{tabular}
\caption{Zero-shot top-5 word prediction accuracy. Top-5 is selected over the whole model vocabulary for the first 4 rows (same as Table \ref{tab:ettinger-word_prediction}). In the last 4 rows, we remove the 179 most common English stop words, as well as the~"~" token from the vocabulary.}
\label{tab:t5-stop-word}
\end{table*}
\endgroup

\begin{table*}[h!]
\centering
\begin{tabular}{ l | c c c c c c c c}
\toprule
& \thead{Age \\  Comp.} & \thead{Always \\  Never} & \thead{Object \\ Comp.} & \thead{Antonym \\ Negation} & \thead{Taxonomy \\ Conj.} & \thead{Multi-hop \\ Comp.} \\
\midrule
        Majority & 50.6 & 36.1 & 50.6 & 50.2 & 34.0 & 34.0 \\
        \midrule
        $\mathrm{BERT}_\mathrm{base}$ & 49.4 $\pm$ 0.2 & 13.2 $\pm$ 1.2 & 55.4 $\pm$ 1.0 & 53.8 $\pm$ 1.0 & 46.8 $\pm$ 0.6 & 33.4 $\pm$ 0.6 \\ 
        $\mathrm{BERT}_\mathrm{large}$ & 50.6 $\pm$ 0.2 & 22.5 $\pm$ 1.3 & 52.4 $\pm$ 1.6 & 50.8 $\pm$ 0.8 & \textbf{53.9 $\pm$ 0.9} & 33.8 $\pm$ 0.7 \\ 
        $\mathrm{BERT}_\mathrm{large}$ WWM & 76.4 $\pm$ 1.7 & 10.7 $\pm$ 1.5 & 55.8 $\pm$ 1.1 & 57.2 $\pm$ 0.7 & 46.4 $\pm$ 0.8 & 33.8 $\pm$ 0.7 \\ 
        $\mathrm{RoBERTa}_\mathrm{large}$ & 98.6 $\pm$ 0.1 & 13.5 $\pm$ 1.6 & 87.4 $\pm$ 0.9 & \textbf{74.6 $\pm$ 0.8} & 45.4 $\pm$ 0.4 & 28.0 $\pm$ 1.0 \\ 
        \midrule
        $\mathrm{DistilBERT}_\mathrm{base}$ & 49.4 $\pm$ 0.2 & 15.0 $\pm$ 1.2 & 51.0 $\pm$ 1.3 & 50.8 $\pm$ 0.7 & 46.8 $\pm$ 0.8 & 34.0 $\pm$ 1.0 \\ 
        $\mathrm{DistilRoBERTa}_\mathrm{base}$ & 45.4 $\pm$ 1.2 & 13.9 $\pm$ 1.3 & 50.8 $\pm$ 0.7 & 51.0 $\pm$ 1.0 & 50.6 $\pm$ 1.1 & 34.0 $\pm$ 1.0 \\ 
        $\mathrm{AlBERT}_\mathrm{base}$ & 47.0 $\pm$ 0.6 & 23.2 $\pm$ 1.2 & 50.6 $\pm$ 0.7 & 52.6 $\pm$ 1.0 & - & 34.0 $\pm$ 1.0 \\ 
        $\mathrm{AlBERT}_\mathrm{large}$ & 52.8 $\pm$ 1.2 & 30.7 $\pm$ 1.0 & 49.2 $\pm$ 0.7 & 50.2 $\pm$ 1.0 & - & 34.0 $\pm$ 1.0 \\ 
        $\mathrm{AlBERT}_\mathrm{xlarge}$ & 39.8 $\pm$ 0.3 & 26.1 $\pm$ 1.5 & 50.4 $\pm$ 0.8 & 44.6 $\pm$ 1.4 & - & 32.2 $\pm$ 1.2 \\ 
        $\mathrm{AlBERT}_\mathrm{xxlarge}$ & 95.4 $\pm$ 0.4 & 22.9 $\pm$ 0.5 & 61.0 $\pm$ 0.7 & 66.4 $\pm$ 0.5 & - & 34.0 $\pm$ 1.0 \\ 
        $\mathrm{AlBERTv2}_\mathrm{base}$ & 50.6 $\pm$ 0.2 & 21.4 $\pm$ 0.9 & 49.4 $\pm$ 0.7 & 54.2 $\pm$ 1.7 & - & 34.0 $\pm$ 1.0 \\ 
        $\mathrm{AlBERTv2}_\mathrm{large}$ & 51.4 $\pm$ 0.6 & 31.7 $\pm$ 1.5 & 50.6 $\pm$ 0.6 & 55.2 $\pm$ 1.3 & - & 34.0 $\pm$ 1.0 \\ 
        $\mathrm{AlBERTv2}_\mathrm{xlarge}$ & 46.2 $\pm$ 0.7 & 37.9 $\pm$ 1.9 & 50.6 $\pm$ 0.7 & 62.4 $\pm$ 0.9 & - & 32.4 $\pm$ 0.8 \\ 
        $\mathrm{AlBERTv2}_\mathrm{xxlarge}$ & 93.8 $\pm$ 0.5 & 23.9 $\pm$ 0.7 & 78.8 $\pm$ 0.8 & 64.8 $\pm$ 0.5 & - & 34.0 $\pm$ 1.0 \\ 
        $\mathrm{BART}_\mathrm{large}$ & 49.4 $\pm$ 0.2 & 23.2 $\pm$ 1.2 & 49.4 $\pm$ 0.7 & 49.8 $\pm$ 1.0 & 48.8 $\pm$ 0.9 & 33.8 $\pm$ 0.7 \\ 
        $\mathrm{T5}_\mathrm{small}$ & 49.4 $\pm$ 0.2 & 16.1 $\pm$ 1.6 & 48.2 $\pm$ 0.8 & 47.0 $\pm$ 0.9 & 49.3 $\pm$ 0.4 & 33.8 $\pm$ 0.7 \\ 
        $\mathrm{T5}_\mathrm{base}$ & 49.4 $\pm$ 0.2 & 10.7 $\pm$ 1.2 & 59.0 $\pm$ 0.7 & 53.4 $\pm$ 0.8 & 46.6 $\pm$ 0.9 & 33.6 $\pm$ 0.7 \\ 
        $\mathrm{T5}_\mathrm{large}$ & 94.0 $\pm$ 0.4 & 25.7 $\pm$ 0.7 & 83.2 $\pm$ 0.5 & 64.6 $\pm$ 1.4 & 42.2 $\pm$ 1.0 & 33.8 $\pm$ 0.7 \\ 
        $\mathrm{T5}_\mathrm{xl}$ & \textbf{100.0 $\pm$ 0.0} & 20.4 $\pm$ 1.0 & \textbf{90.0 $\pm$ 0.5} & 68.4 $\pm$ 0.8 & 41.2 $\pm$ 0.8 & 34.4 $\pm$ 0.6 \\ 
        $\mathrm{T5v1.1}_\mathrm{small}$ & 49.4 $\pm$ 0.2 & 34.3 $\pm$ 1.8 & 50.6 $\pm$ 0.7 & 51.4 $\pm$ 1.1 & 48.2 $\pm$ 0.7 & \textbf{37.8 $\pm$ 0.9} \\ 
        $\mathrm{T5v1.1}_\mathrm{base}$ & 50.6 $\pm$ 0.2 & 11.8 $\pm$ 1.6 & 56.0 $\pm$ 1.5 & 45.0 $\pm$ 0.8 & 49.9 $\pm$ 0.7 & 37.6 $\pm$ 0.9 \\ 
        $\mathrm{T5v1.1}_\mathrm{large}$ & 49.6 $\pm$ 0.3 & 15.7 $\pm$ 0.8 & 50.6 $\pm$ 0.8 & 47.1 $\pm$ 1.1 & 41.7 $\pm$ 1.0 & 33.8 $\pm$ 0.7 \\ 
        $\mathrm{T5v1.1}_\mathrm{xl}$ & 49.4 $\pm$ 0.2 & 23.9 $\pm$ 1.8 & 49.4 $\pm$ 0.7 & 54.2 $\pm$ 1.2 & \textbf{53.9 $\pm$ 0.5} & 33.8 $\pm$ 0.7 \\ 
        $\mathrm{UniLM}_\mathrm{base}$ & 47.9\textpm1.6 & 16.1\textpm0.8 & 48.0\textpm2.7 & 43.6\textpm1.3 & 45.1\textpm1.2 & 34.8\textpm0.9 \\ 
        $\mathrm{UniLM}_\mathrm{large}$ & 47.9\textpm1.6 & 19.9\textpm1.3 & 61.4\textpm1.8 & 51.2\textpm1.4 & 50.2\textpm2.1 & 33.6\textpm0.7 \\
        $\mathrm{GPT2}_\mathrm{base-0.1B}$ & 47.6\textpm1.2 & \textbf{50.1\textpm1.5} & 50.1\textpm1 & 52.8\textpm1.9 &  48.4\textpm1.0 & 32.2\textpm2.4 \\
        $\mathrm{GPT2}_\mathrm{medium-0.3B}$ & 50.1\textpm1.3 & 40.8\textpm2.2 & 49.6\textpm0.9 & 54.7\textpm2.4 &  49.1\textpm1.7 & 29.6\textpm2.1 \\ 
        $\mathrm{GPT2}_\mathrm{large-0.8B}$ & 69.6\textpm1.0 & 20.2\textpm1.7 & 50.4\textpm1.0 & 50.1\textpm2.7 &  46.9\textpm1.5 & 33.5\textpm1.3 \\ 
        $\mathrm{GPT}_\mathrm{NEO-1.3B}$ & 58.6\textpm0.7 & 29.0\textpm1.0 & 52.1\textpm0.7 & 65.2\textpm1.1 & 50.6\textpm1.5 & 33.3\textpm1.0 \\
        $\mathrm{GPT2}_\mathrm{xl-1.5B}$ & 51.9\textpm1.5 & 26.6\textpm0.7 & 52.6\textpm0.7 & 60.6\textpm1.2 & 45.8\textpm1.3 & 34.0\textpm1.0 \\
        \bottomrule
\end{tabular}
\caption{Zero-shot oLMpics evaluation on MC-MLM tasks. ``Majority'' here is the accuracy when predicting the most frequent class. The first 4 models are our reproduction of the original oLMpics results.
The best result on each task is highlighted in bold.
Confidence intervals estimated via bootstrapping 20\% of the data show errors about 1-2 absolute points.
}
\label{tab:mc-mlm-zero-shot-conf-intervals}
\end{table*}

\end{document}